\documentclass{article}
\usepackage{spconf,amsmath,graphicx}
\usepackage[utf8]{inputenc} 
\usepackage[T1]{fontenc}    
\usepackage{hyperref}       
\usepackage{url}            
\usepackage{booktabs}       
\usepackage{amsfonts}       
\usepackage{nicefrac}       
\usepackage{microtype}      
\usepackage{xcolor}         

\usepackage{multirow}
\usepackage{graphicx}
\usepackage{colortbl}
\usepackage{algorithm}
\usepackage{algorithmic}
\usepackage{amsmath}
\usepackage{wrapfig}
\usepackage{enumitem}
\usepackage{marvosym}
\usepackage{subfigure}

\title{Label-Occurrence-Balanced Mixup for long-tailed recognition}
\name{\begin{tabular}{c}Shaoyu Zhang$^{1,2}$, Chen Chen$^{1,2*}$, Xiujuan Zhang$^{3}$, Silong Peng$^{1,2}$
\end{tabular}\thanks{$^*$ Corresponding author.}}
\address{
    $^{1}$Institute of Automation, Chinese Academy of Sciences, China\\
    $^{2}$ University of Chinese Academy of Sciences, China\\
    $^{3}$ Inner Mongolia Key Laboratory of Molecular Biology on Featured Plants, China\\
    }
\begin{document}
%
\maketitle
\begin{abstract}
Mixup is a popular data augmentation method, 
with many variants subsequently proposed. 
These methods mainly create new examples via convex combination of random data pairs 
and their corresponding one-hot labels.
However, most of them adhere to a random sampling and mixing strategy, 
without considering the frequency of label occurrence in the mixing process. 
When applying mixup to long-tailed data, a {\itshape label suppression} issue arises, 
where the frequency of label occurrence for each class is imbalanced 
and most of the new examples will be completely or partially
assigned with head labels. 
The suppression effect may further aggravate the problem of data imbalance 
and lead to a poor performance on tail classes.
To address this problem, we propose {\itshape Label-Occurrence-Balanced Mixup} to augment data while
keeping the label occurrence for each class statistically balanced. 
In a word, we employ two independent class-balanced samplers to select data pairs 
and mix them to generate new data. 
We test our method on several long-tailed vision and sound recognition benchmarks. 
Experimental results show that our method significantly promotes the adaptability of mixup method 
to imbalanced data and achieves superior performance compared with state-of-the-art
long-tailed learning methods.

\end{abstract}
\begin{keywords}
Long-tailed learning, mixup, data augmentation, class-balanced sampler, vision and sound recognition
\end{keywords}
\section{Introduction}
\label{sec:intro}

Deep convolutional neural networks 
have led to a series of breakthroughs for visual and sound 
recognition. 
The training of such networks often needs a rich supply of data to 
improve the generalization ability.
In this regard, a number of data augmentation and regularization techniques
have been recently proposed, 
including mixup-based methods \cite{zhang2018mixup, tokozume2018learning}.
\begin{figure}[h]
  \centering
    \includegraphics[width=\linewidth]{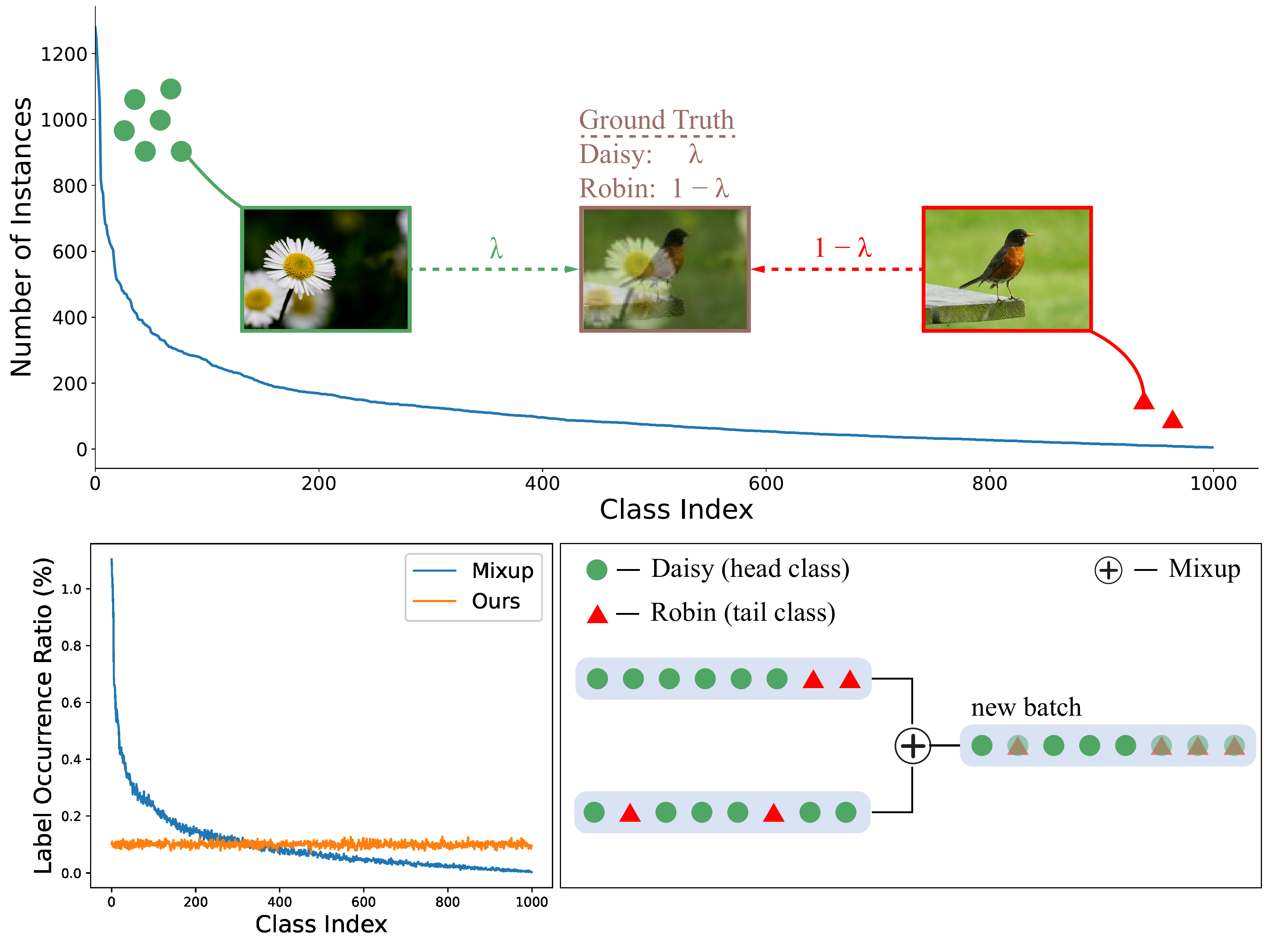}
  \vspace{-0.1cm}
  \caption{
  {\itshape Top}: Illustration of the number of instances per class in ImageNet-LT, 
  the distribution of which follows a long-tailed distribution. 
  Mixup creates examples by convex combinations of data pairs and their labels.
  {\itshape Bottom}: Applying mixup to long-tailed data leads to {\itshape label suppression}, 
  where head labels occupy the main position and most of the created data will be assigned with 
  head labels ({\itshape right}). 
  Label occurrence ratio for each class is highly imbalanced in 
  the mixing process, while our method 
  re-balance this ratio among classes ({\itshape left}).}
  \label{fig:intro}
\end{figure}

Mixup \cite{zhang2018mixup}, 
which generates new examples by combining random data pairs and their labels, 
has shown promising performance on 
model generalization \cite{zhang2020does} and calibration \cite{thulasidasan2019mixup}. 
Motivated by this idea, many follow-up methods 
\cite{yun2019cutmix,verma2019manifold} 
were proposed and have proved to be effective on commonly used datasets, 
e.g., CIFAR and ImageNet ILSVRC 2012 \cite{deng2009imagenet}.
However, these datasets are often collected with relatively balanced data distribution among classes. 
In contrast, real-world data often follow a 
long-tailed distribution, 
where head classes occupy a significantly larger number of data than tail classes.
When handling such imbalanced data, 
mixup may not be an effective method to improve performance. 
Instead, directly mixing random data pairs and labels may 
cause a problem
that label occurrence among classes is imbalanced and
most of the mixed examples will be embedded with 
features and labels from head classes, which we called {\itshape label suppression}.

In this paper, we introduce the concept of {\itshape label occurrence ratio} to demonstrate 
the phenomenon of label suppression. 
As training data pairs and their labels are mixed with random mixing ratios, 
each new example may belong to multiple classes proportionally. 
After mixup, the expected volume of a class in new data can be represented by 
the sum of the mixing ratios from this class. 
We therefore define the {\itshape label occurrence ratio} as the proportion of 
the expected volume of a class among all newly created data. 
As shown in Fig. \ref{fig:intro}, 
the label occurrence ratios for different classes 
from ImageNet-LT are highly imbalanced after mixup. 
In addition, about $94.7\%$ of new examples will be 
completely or partially assigned with head labels. 
The suppression effect actually introduces noise to tail data and 
further increases difficulty in learning tail classes.

To address this problem, a natural idea is to balance the label occurrence, 
either by applying class-conditional mixing ratio or class-conditional sampling. 
Considering that the former may lead to very small mixing ratio for head classes 
and fail to combine informative features, 
we apply the latter and adjust sampling strategy to re-balance the distribution of label occurrence.
In this regard, class-balanced sampling \cite{shen2016relay}
is a commonly used re-balancing method to learn imbalanced data.
Although effective, recent study \cite{cao2019learning, zhou2020bbn, kang2019decoupling}
find it may lead to overfitting 
on tail classes and hurt the representation learning. 
However, in long-tailed scenarios, 
we observe a natural complementarity of class-balanced sampling method 
and mixup method: mixup method increases the diversity of sampled data and alleviates risk
of overfitting on tail classes, while class-balanced sampling helps to keep the mixed 
label occurrence relatively balanced to alleviate label suppression and 
learn unbiased classifier. 
Motivated by this observation, we propose label-occurrence-balanced mixup, 
which employs two independent class-balanced samplers to generate data pairs 
with balanced label occurrence among classes (see bottom left of Fig. \ref{fig:intro}), 
and then mixes the data pairs to create new data.
Despite its simplicity, our method effectively
generalizes the mixup-based methods to real-world long-tailed data. 

The main contributions of this paper are as follows: 
\begin{itemize}
    \item We explore mixup-based methods in long-tailed scenarios and
    analyze their failure to achieve expected performance due to label suppression. 
    We further define the label occurrence ratio to demonstrate this phenomenon.
    \item We discuss the merits/demerits of mixup and class-balanced sampling, 
    and discover a complementary property of these two methods.
    \item We propose 
    label-occurrence-balanced mixup to alleviate label suppression 
    and significantly improve the performance of mixup in long-tailed scenarios. 
\end{itemize}

\section{Related Woks}
\label{sec:related}

\noindent
{\bfseries Mixup.}\quad 
Mixup training \cite{zhang2018mixup}, 
which shares the same idea with between-class learning in sound recognition \cite{tokozume2018learning}, 
has been shown to substantially improve model generalization and robustness \cite{zhang2020does}. 
Manifold mixup \cite{verma2019manifold} extends the idea of mixing data pairs
from input space to feature space. 
Recently, Yun et al. \cite{yun2019cutmix} propose CutMix via regional replacement of random data pairs.
Besides supervised settings, mixup-based methods also prove to be effective 
in semi-supervised learning \cite{verma2019interpolation}. 

\noindent
{\bfseries Long-tailed Recognition.}\quad 
Recent advances in tackling long-tailed challenges are mainly based on re-balancing methods 
and meta-learning methods. 
Re-balancing methods can be divided into two regimes: 
re-sampling \cite{buda2018systematic} and 
cost-sensitive re-weighting \cite{cui2019class, cao2019learning}.
Re-sampling methods create a relatively balanced data distribution by over-sampling, under-sampling, 
or class-balanced sampling \cite{shen2016relay}, 
while re-weighting methods design class-wise weights to adjust learning
focus on different classes. 
In addition, some meta-learning based methods \cite{liu2019large}
facilitates learning tail data by 
transferring the information from head classes to tail classes. 
Beyond that, Remix \cite{chou2020remix}, which applies mixup to long-tailed scenarios, 
adjusts label distribution by mixing 
some head-tail data pairs while keeping the ground truth only being the tail label. 
However, this strategy does not guarantee a balanced label distribution and 
meantime introduces noise to tail classes.

\section{Method}
\label{sec:method}

In this section, we firstly provide an overview of mixup method 
and introduce a metric for evaluating label suppression, 
and then describe the proposed label-occurrence-balanced mixup to alleviate the problem.

\subsection{Ovreview of Mixup Method}
\label{ssec:overview}
Instead of Empirical Risk Minimization (ERM), mixup method 
creates new examples in the vicinity of original data and trains model based 
on the principle of Vicinal Risk Minimization (VRM) \cite{chapelle2001vicinal}. 
To be specific, given training data $x$ and its label $y$, 
mixup creates new example $(\tilde{x}, \tilde{y})$ 
by combining two random training data $(x_i, y_i)$ and $(x_j, y_j)$ linearly
\begin{gather}
    \tilde{x} = \lambda x_i + (1-\lambda) x_j \\
    \tilde{y} = \lambda y_i + (1-\lambda) y_j,
\end{gather}
the combination ratio $\lambda \in (0,1)$ between the two data points is 
sampled from the beta distribution ${\textup{Beta}}(\alpha,\alpha)$. 
\begin{figure}[h]
  \centering
    \includegraphics[width=\linewidth]{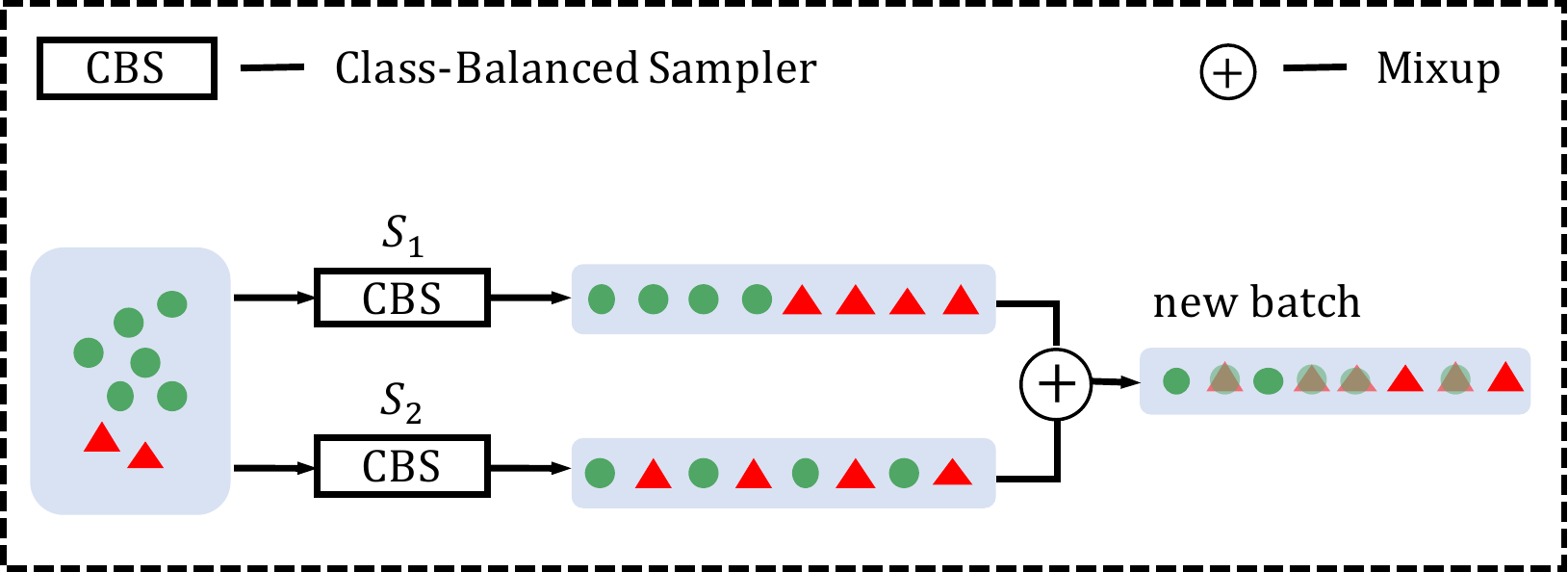}
  \caption{
  Framework of our label-occurrence-balanced mixup. 
  We apply two class-balanced samplers, where head classes (green) and tail classes (red)
  have an equal probability of been selected. 
  Each of the sampler works independently and 
  generates a data batch with uniform distribution among classes.
  Then a new batch with balanced label occurrence is created by mixing the two batches.}
  \label{fig:lob}
\end{figure}

Due to the randomness in selecting data pairs and combination ratio, 
mixup relaxes the constraint of finite training datasets and 
encourages model to learn on diverse in-between examples. 
However, the randomness in sampling data will lead to an imbalanced class distribution 
for long-tailed data.

\subsection{Label-Occurrence-Balanced Mixup}
\label{ssec:lob}
Here, we first introduce {\itshape label occurrence ratio} to quantitatively demonstrate the phenomenon 
of label suppression. 
Formally, 
for a dataset consisting of $N$ data points from $C$ classes, 
mixup shuffles the dataset twice and 
mixes the $2N$ data points to generate $N$ new examples. 
For convenience, we double the index and let $A\equiv \{1, \dots, 2N\}$ 
be the set of indexes of data points from the two shuffles and 
$I(k)\equiv \{i\mid y_i=k, i\in A \}$ be the set of indexes of data points from class $k$. 
In addition, the number of data points in class $k$ is denoted by $n_k=\left| I(k) \right|$.
Assuming the probability of data point $(x_i, y_i)$ being selected is $P_i$, 
we define the {\itshape label occurrence ratio} for class $k$ as 
\begin{equation}\label{lor}
    \gamma_k = \frac{\sum_{i\in I(k)}P_i\lambda_i}{\sum_{j\in A}P_j\lambda_j},
\end{equation}
where $\lambda_i$ is the corresponding mixing ratio of $(x_i, y_i)$. 
In mixup, each data point has the same probability of being selected, 
i.e., $P_i=1/(2N)$ for $i\in A$. 
Because the number of data points $n_k$ for each class is imbalanced, 
the distribution of $\gamma_k$ is also imbalanced, 
which leads to label suppression. 

To address the problem of label suppression, 
the key idea is to balance the distribution of $\gamma_k$, 
either by adjusting $\lambda_i$ or $P_i$. 
We find the former method of adjusting $\lambda_i$ is sub-optimal, 
as the mixing ratio for all the head examples will be too small to provide informative feature.
Therefore, we re-balance the label occurrence by adjusting $P_i$. 
As shown in Fig. \ref{fig:lob}, 
we employ two independent class-balanced samplers $S_1$ and $S_2$ to generate data pairs. 
In this case, 
each class has an equal probability of being selected by the two samplers. 
The probability of sampling an example with label $k$ is
\begin{equation}
    p_k = \frac{1}{C}.
\end{equation}
Accordingly, the probability of an example $(x_i, y_i)$ of being selected is class-conditional: 
\begin{equation}
    P_{i\mid i\in I(k)} = \frac{1}{n_k}p_k = \frac{1}{n_kC}.
\end{equation}
The above process can be seen as a two-stage sampling operation 
from one class list and $C$ per-class sample lists, 
which first selects a class $k$ from the class list 
and then samples an example from the per-class sample list
of class $k$ uniformly. 

During training, the samplers generate two data points, 
$(x_i^{S_1}, y_i^{S_1})$ from $S_1$ and $(x_j^{S_2}, y_j^{S_2})$ from $S_2$, respectively.
For exhaustively taking advantage of the data diversity, 
both of the data points are sampled from the whole dataset independently, 
which are not limited in the same mini-batch. 
Then we perform mixup on the data pair to create new examples:
\begin{gather}
    \tilde{x}_{\textup {LOB}} = \lambda x_i^{S_1} + (1-\lambda) x_j^{S_2} \\
    \tilde{y}_{\textup {LOB}} = \lambda y_i^{S_1} + (1-\lambda) y_j^{S_2}.
\end{gather}
Due to the dual class-balanced samplers, the new batch composed of 
$(\tilde{x}_{\textup {LOB}}, \tilde{y}_{\textup {LOB}})$ have a balanced 
distribution of $\gamma_k$.

To further improve the training performance, 
we employ a deferred re-balancing training strategy \cite{cao2019learning}, 
which first trains with vanilla mixup before annealing the learning rate, 
and then uses the proposed label-occurrence-balanced mixup.

\section{Experiments}
\label{sec:exp}

We conduct experiments on four long-tailed visual and sound recognition benchmarks 
and different backbone networks to prove the effectiveness of the proposed method.
\subsection{Datasets}
\label{ssec:datasets}

\noindent
{\bfseries ESC-50-LT.}\quad 
ESC-50 \cite{piczak2015esc} contains a total of 2000 environmental recordings equally 
balanced between 50 classes. 
We select 8 examples per class to form validation set, 
and sample the rest of examples following Pareto distribution to form a long-tailed training set.
The imbalance ratio $\rho$ denotes 
the ratio between the number of examples of
the most frequent class and the least frequent class. We set $\rho=10$ 
for ESC-50-LT.

\noindent
{\bfseries CIFAR-10-LT and CIFAR-100-LT.}\quad 
Following the prior work \cite{cui2019class, cao2019learning}, 
we use the long-tailed
version of the CIFAR-10 and CIFAR-100 datasets with $\rho=10,100$. 

\noindent
{\bfseries ImageNet-LT.}\quad 
ImageNet-LT is constructed by sampling a long-tailed subset of ImageNet-2012 \cite{deng2009imagenet}. 
It has 115.8K images
from 1000 categories, with the number of images per class ranging from 1280 to 5. 

\subsection{Implementation Settings}
\label{ssec:implementation} 
For sound recognition on ESC-50-LT, we use EnvNet \cite{tokozume2017learning} 
and EnvNet-v2 \cite{tokozume2018learning} as backnone networks 
and follow the training settings of \cite{tokozume2018learning}. For visual recognition on 
CIFAR-10/CIFAR-100-LT, we train a ResNet-32 backnone network for 200 epochs, 
with a learning rate initialized as 0.1 and decayed at the 160th and 180th epoch. 
For ImageNet-LT, we choose ResNet-10 as the backbone network and train the model for 90 epochs, 
following \cite{kang2019decoupling}. 
The base learning rate is set to 0.2, with cosine learning rate decay. 
The mixing ratio $\lambda \sim {\textup{Beta}}(\alpha,\alpha)$, where we set $\alpha=0.2$ for ImageNet-LT,
and $\alpha=1$ for other datasets. 

\begin{table}[htbp]
	\centering	
	\resizebox{\columnwidth}{!}{
	\begin{tabular}{|l|cc|cc|c|}		
		\hline		
		Dataset & \multicolumn{2}{c|}{CIFAR-10-LT} & \multicolumn{2}{c|}{CIFAR-100-LT} &
		\multicolumn{1}{c|}{ImageNet-LT}\\	
        \hline
		Imbalance ratio &  100   & 10 &  100   &  10  &  256 \\  
		\hline
		ERM  &  29.7  &  12.9  &  60.7  &  43.4  &  65.4\\
		\hline
		Mixup \cite{zhang2018mixup}  &  28.4  &  11.5  &  59.1  &  41.6  &  67.2\\ 
		Manifold Mixup \cite{verma2019manifold}  &  30.2  &  13.3  &  60.4  &  42.6  &  67.5\\ 
		Remix \cite{chou2020remix}  &  27.0  &  11.5  &  58.6  &  40.5  &  66.6\\ 
		Ours  &  25.8  &  10.6  &  58.5  &  40.1  &  63.0\\
		\hline
		CB Samp. \cite{shen2016relay}  &  31.6  &  13.1  &  68.1  &  45.0  &  64.4\\
		CB Samp.* \cite{cao2019learning}  &  26.5  &  12.3  &  58.5  &  42.4  &  60.1\\
		\hline
		LDAM-DRW \cite{cao2019learning}  &  23.0  &  11.8  &  58.0  &  41.3  &  64.0\\
        BBN \cite{zhou2020bbn}  &  22.0  &  12.7  &  57.4  &  40.9  &  -\\
		Logit Adj. \cite{50403}  &  22.3  &  11.8  &  56.1  &  42.3  &  -\\
		LFME \cite{xiang2020learning}  &  -  &  -  &  57.7  &  -  &  62.8\\
		\hline
		Ours*  &  {\bfseries 21.3}&  {\bfseries 10.4}    &  {\bfseries 53.8}&  {\bfseries 38.9}  &
		{\bfseries 59.6}\\
		\hline		
	\end{tabular}	
	}
	\vspace{-0.2cm}
	\caption{Top-1 validation error rates on CIFAR-10-LT/CIFAR-100-LT and ImageNet-LT. 
	$*$ denotes the deferred re-balancing version of corresponding methods.}
	\label{visual}
\end{table}

\begin{table}[htbp]
	\centering	
	\resizebox{\columnwidth}{!}{
	\begin{tabular}{|l|cc|cc|}		
		\hline		
		Backbone & \multicolumn{2}{c|}{EnvNet \cite{tokozume2017learning}} &
		\multicolumn{2}{c|}{EnvNet-v2 \cite{tokozume2018learning}}\\	
        \hline
		Error rates $(\%)$ &  Top-1 error   & Top-5 error &  Top-1 error   &  Top-5 error \\  
		\hline
		ERM  &  54.7  &  24.2  &  52.2  &  23.7\\
		CB Samp. \cite{shen2016relay}  &  55.7  &  25.3  &  54.6  &  29.3\\
		BC{\scriptsize{(mixup)}} \cite{tokozume2018learning}  &  47.0  &  25.2  &  45.2  &  20.8\\ 
		\hline
		Ours  &  {\bfseries 44.9}  &  {\bfseries 21.4}  &  {\bfseries 42.6}  &  {\bfseries 19.2}\\
		\hline		
	\end{tabular}	
	}
	\vspace{-0.2cm}
	\caption{Top-1/Top-5 validation error rates on ESC-50-LT for EnvNet and EnvNet-v2.}
	\label{sound}
\end{table}

\subsection{Experimental Results}
\label{ssec:results}
\noindent
{\bfseries Competing methods.}\quad 
We compare the proposed label-occurrence-balanced mixup 
with ERM baseline and three groups of methods:
1) mixup-based methods, including mixup training \cite{zhang2018mixup}, 
manifold mixup \cite{verma2019manifold} and Remix \cite{chou2020remix}, 
where mixup is replaced by between-class (BC) learning 
\cite{tokozume2018learning} for sound recognition on ESC-50-LT; 
2) sampling-based methods, including class-balanced sampling (CB Samp.) \cite{shen2016relay}, 
deferred class-balanced sampling (CB Samp.*) \cite{cao2019learning}; 
3) state-of-the-art methods, including recently proposed LDAM-DRW \cite{cao2019learning}, 
BBN \cite{zhou2020bbn}, logit adjustment loss \cite{50403} and LFME \cite{xiang2020learning}. 
Our method is denoted as {\itshape Ours}, and the deferred re-balancing version of our method 
is denoted as {\itshape Ours*}.

\begin{figure}[htb]

\begin{minipage}[b]{1.0\linewidth}
  \centering
  \centerline{\includegraphics[width=8.5cm]{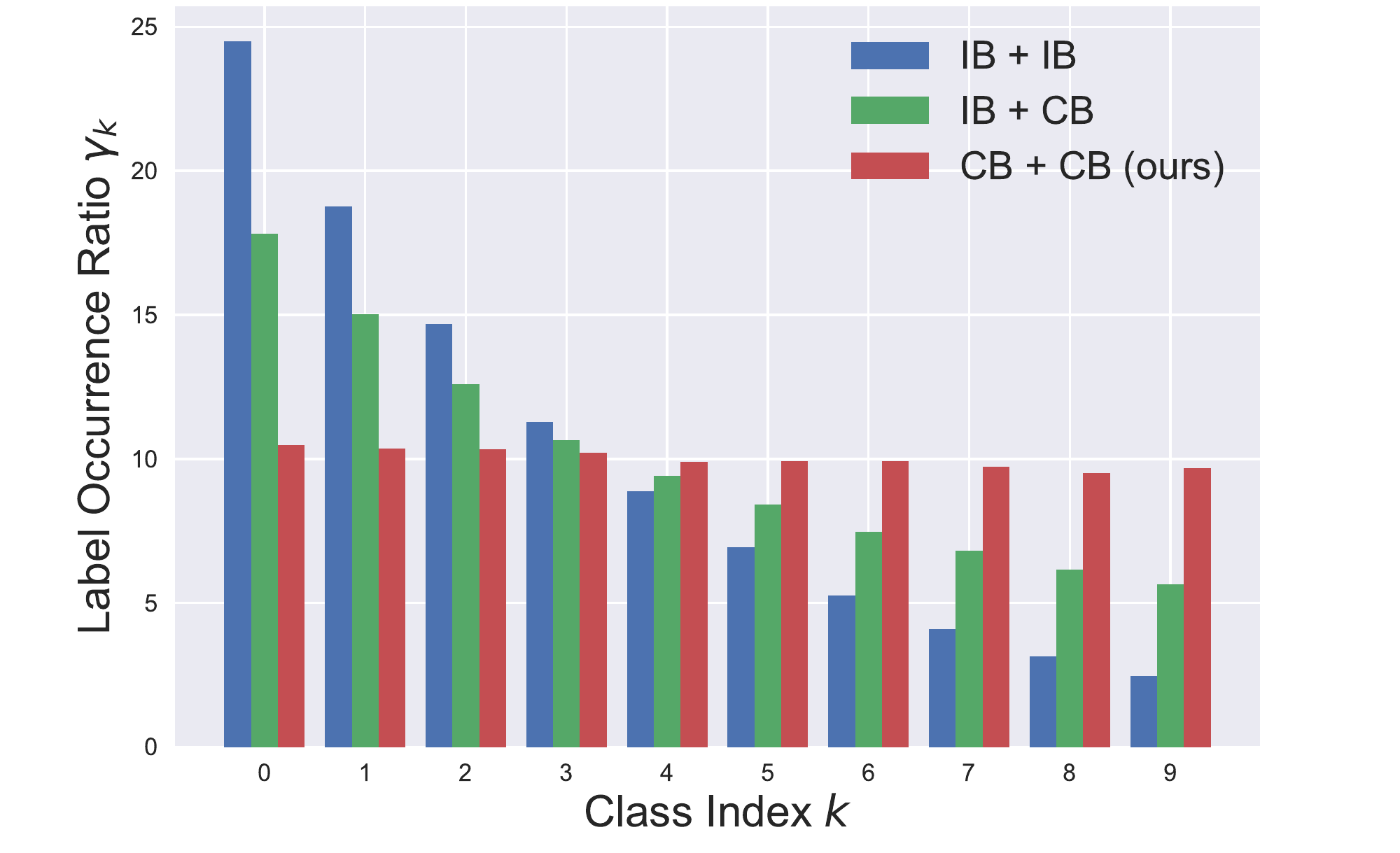}}
  \centerline{(a) $\gamma_k$ for alternating samplers on CIFAR-10-LT, $\rho=10$.}\medskip
\end{minipage}
\hfill
\begin{minipage}[b]{1.0\linewidth}
\centering
\resizebox{\linewidth}{!}{
\begin{tabular}{|l|c|c|}		
        \hline
		Combination of samplers &  $\gamma_{\max}/\gamma_{\min}$   & Top-1 error $(\%)$  \\  
		\hline
		IB Sampler + IB Sampler (mixup)  &  9.91  &  11.5\\
		IB Sampler + CB Sampler  &  3.16  &  11.4\\ 
		CB Sampler + CB Sampler (ours)  &  {\bfseries 1.10}  &  {\bfseries 10.6}\\
		\hline		
	\end{tabular}	
	}
    ~\\
    \centerline{(b) The correlation between performance and balance of $\gamma_k$.}\medskip
\end{minipage}
\vspace{-0.7cm}
\caption{Ablation study for alternating the two samplers on CIFAR-10-LT, with imbalance ratio $\rho=10$.}
\label{ablation}
\end{figure}

Results for visual recognition are reported in Table \ref{visual}. 
Compared with mixup, our method obtains $4.2\%$ relative improvement on ImageNet-LT. 
Our method also outperforms other mixup-based methods like manifold mixup and Remix.
It is worth noting that class-balanced
sampling method leads to even worse performance on some datasets, e.g., CIFAR-100-LT, while 
our method shows consistent performance gains on all the reported benchmarks. 
Furthermore, by integrating with deferred re-balancing training strategy, 
our method achieves lower error rates than most of the state-of-the-art methods, 
such as logit adjustment \cite{50403} and BBN \cite{zhou2020bbn}. 

Results for sound recognition show similar trends. 
As shown in Table \ref{sound}, class-balanced sampling gets worse performance, 
probably due to overfitting on repeated samples.
BC learning outperforms ERM, 
while our method further improves the performance for both EnvNet and EnvNet-v2 backbones.

\subsection{Ablation Study}
\label{ssec:ablation}

\noindent
{\bfseries Alternating two samplers.}\quad 
The key idea of our method is to balance the label occurrence ratio by 
employing two class-balanced (CB) samplers. 
Here we discuss the effect of alternating the two samplers. 
Mixup could be seen as using two instance-balanced (IB) samplers, 
where each example has the same probability of being selected. 
Beyond that, there is another case that an instance-balanced sampler and a class-balanced sampler 
are both used.  
We analyze the correlation between model performance and the balance of $\gamma_k$. 
Fig. \ref{ablation}(a) shows the distribution of $\gamma_k$ for the three cases 
of combining samplers. We find that both the cases of IB Sampler + IB Sampler and 
IB Sampler + CB Sampler lead to an imbalanced distribution of $\gamma_k$, 
while our method achieves a balanced distribution. 
From Fig. \ref{ablation}(b) we can see that a more balanced $\gamma_k$, i.e.,  
$(\gamma_{\max}/\gamma_{\min}\to 1)$, 
leads to a better model performance.

\noindent
{\bfseries Adjusting $\lambda$ or adjusting $P$.}\quad 
In Equation \ref{lor}, $\gamma_k$ could be adjusted either by adjusting $\lambda$ or $P$. 
For the former, to balance the distribution of $\gamma_k$,
one may reduce each mixing ratio $\lambda$ for head examples to a very small value, 
which is impracticable to provide informative features.
Instead of constraining the mixing ratio $\lambda$ directly, 
our method balances the summation of $\lambda$ by controlling the 
probability $P$ of sampling example from different classes.
In experiments, adjusting $\lambda$ achieves top-1 error rates of $26.2\%$ and $15.9\%$ for 
CIFAR-10-LT with $\rho=100$ and $\rho=10$, respectively, 
while our method achieves $25.8\%$ and $10.6\%$, which shows more robust performance.

\section{Conclusion}
\label{sec:conclusion}

In this paper, we propose label-occurrence-balanced mixup, 
which addresses the problem of label suppression and
generalizes mixup-based methods to real-world long-tailed scenarios. 
Label-occurrence-balanced mixup is a simple and effective method that
shows consistent improvements on several vision and sound 
recognition benchmarks.

\vfill\pagebreak

\bibliographystyle{IEEEbib}
\bibliography{refs}

\end{document}